\documentclass{article}

     \usepackage[preprint]{neurips_2020}

\usepackage[utf8]{inputenc} 
\usepackage[T1]{fontenc}    
\usepackage{hyperref}       
\usepackage{url}            
\usepackage{booktabs}       
\usepackage{amsfonts}       
\usepackage{nicefrac}       
\usepackage{microtype}      

\usepackage{mathrsfs}
\usepackage{tikz}
\usepackage{pgfplots}
\usepackage{tabularx}
\usetikzlibrary{positioning,matrix}
\usepackage{stmaryrd}

\usepackage{graphicx}
\usepackage{subfigure}

\title{Edge-Featured Graph Attention Network}

\author{%
  Jun Chen\\
  School of Software\\
  Shanghai Jiao Tong University\\
  Shanghai, China \\
  \texttt{thunderboy@sjtu.edu.cn} \\
  
  \And
  
  Haopeng Chen \\
  School of Software\\
  Shanghai Jiao Tong University\\
  Shanghai, China \\
  \texttt{chen-hp@sjtu.edu.cn} \\
}

\begin{document}

\maketitle

\begin{abstract}
  Lots of neural network architectures have been proposed to deal with learning tasks on graph-structured data. However, most of these models concentrate on only node features during the learning process. The edge features, which usually play a similarly important role as the nodes, are often ignored or simplified by these models. In this paper, we present edge-featured graph attention networks, namely EGATs, to extend the use of graph neural networks to those tasks learning on graphs with both node and edge features. These models can be regarded as extensions of graph attention networks (GATs). By reforming the model structure and the learning process, the new models can accept node and edge features as inputs, incorporate the edge information into feature representations, and iterate both node and edge features in a parallel but mutual way. The results demonstrate that our work is highly competitive against other node classification approaches, and can be well applied in edge-featured graph learning tasks.
\end{abstract}

\section{Introduction}

In many real-world applications, data are best constructed as graphs to analyze and display. The graph is such a natural structure whose nodes and edges can be used to characterize the entities and their inner-relationships among data. Recently, several works have defined neural networks on graphs\cite{zhou2018graph, zhang2018deep}. Kipf et al.\cite{kipf2016semi} proposed graph convolutional networks, namely GCNs, based on spectral graph theory. Veli{\v{c}}kovi{\'c} et al.\cite{velivckovic2017graph} presented graph attention networks (GATs), which aggregate features following a self-attention strategy. Though such graph neural networks have been proven successful in some of node classification tasks, they still have obvious shortcomings. One of these networks' major problems is that the edge features are not incorporated into the models. In fact, most of the current state-of-the-art graph neural networks have not consider the edge features.

However, edges with their features play an essential role in many real-world node classification tasks. For example, in a trading network, the node labels may be highly relevant to the transactions. In such a case, the information contained in edges may have a more significant contribution to the classification accuracy compared with node features. Actually, different graphs have different preferences for features of nodes and edges, whereas all existing GNNs ignore or shun this fact.

In this paper, we proposed edge-featured graph attention networks (EGATs) to address the above challenges. This work can be regarded as an extension of GATs. To exploit the edge features effectively, we enhance the original attention mechanism; thus, the edge information can be an important factor in attention-weight computing. Further, the structures and learning processes of traditional attention models are also redesigned in our work, so the models can accept both node and edge features and iterate them individually. The updating of edge features is necessary and should be node-equivalent, because an iterative consistency between nodes and edges should be kept during the learning. Besides, a multi-scale merge strategy, which concatenates features from different iterations, is also adopted in our work. All node and edge features will be gathered in the final layer so that the model can learn the necessary characteristics which benefit the classification from various scales. 

To our best knowledge, we are the first to incorporate edges into GATs as equivalent entities like nodes, point out graphs' different different preferences for features, and handle them spontaneously within the models. Our models can be applied to graphs with discrete and continuous features for both nodes and edges, which can satisfy the demands of many real-world node classification tasks.

\section{Related work}

Graph neural networks (GNNs)\cite{scarselli2008graph} first extended neural network architectures to graphs. As a result, many related works were sprung up ensuingly. Spectral approaches, established on spectral graph theory, are among the most critical parts of these works. Bruna et al.\cite{bruna2013spectral} first defined convolution operations on Fourier domain. However, the filters this model computed are non-spatially localized, so Henaff et al.\cite{henaff2015deep} improved it by generating spatially localized filters. Kipf et al.\cite{kipf2016semi} proposed graph convolutional networks (GCNs), which further simplified above methods by using a first-order approximation on the Chebyshev polynomials. Unlike GCNs, Veli{\v{c}}kovi{\'c} et al.\cite{velivckovic2017graph} proposed graph attention networks (GATs) to dynamically aggregate node features. Numerous variants have been derived from such design. Wang et al.\cite{wang2019heterogeneous} introduced heterogeneous graph attention networks (HANs) to process various heterogeneous graphs. Ma et al.\cite{ma2019disentangled} put forward disentangled graph convolutional networks using a routing algorithm. Several other works also made efforts to learn graph representations. GraphSAGE\cite{hamilton2017inductive} generates node embeddings by aggregating node features using several pre-defined aggregate operations. Inspired by RNNs like LSTM\cite{hochreiter1997long} and GRU\cite{cho2014learning}, gated graph neural networks\cite{li2015gated}, namely GGNNs, were proposed. Furthermore, Xu et al.\cite{xu2018representation} explored jumping knowledge networks, in which layer aggregation is adopted to acquire multi-scale features.

However, all these graph neural networks have a common characteristic: focus on node rather than edge features. Only very few works have tried to integrate edge features into GNN architecture. Schlichtkrull et al.\cite{schlichtkrull2018modeling} proposed an extension architecture of GCNs named R-GCNs. Gong et al.\cite{gong2019exploiting} presented a framework that augments GCNs and GATs with edges. However, such approaches are somewhat not as reasonable as they are. We will highlight their limitations in the next section.

\section{Motivation}

The demands of processing edge-featured graphs are quite common in real-world tasks. For example, if there is a need to find users who may have illegal behaviors in a trading network, it is better to use some node classification approaches to pick them up. Evidently, one user is suspicious or not is highly relevant to the amount he paid or received some time. In other words, the edge features are likely to have a more significant impact on classification than node features under such a situation. However, traditional GNNs cannot handle these graphs in a direct, elegant, and reasonable way. There may be some doubts that if it is possible to convert graphs to which current models can readily accept. Obviously, ignoring edge features is unacceptable. Using pre-defined aggregate functions to integrate edge features into nodes may be a better solution. We do not deny it may perform well on certain graphs, whereas it is not a panacea suitable for every condition, for the selection of function is highly dependent on graphs' traits. It is more like feature engineering, rather than a universal approach.

Only a few works exploit edge features in graph neural networks, and all of them have obvious limitations. Schlichtkrull et al.\cite{schlichtkrull2018modeling} proposed R-GCNs to process modeling relational data. However, the models can accept graphs only when their edges are labeled, which indicates that the edges cannot include continuous attributes. Gong et al.\cite{gong2019exploiting} presented a framework enhances GCNs\cite{kipf2016semi} and GATs\cite{velivckovic2017graph}. This framework can accept continuous attributes of edges, whereas it merely regards them as weights between different node pairs. In most cases, it is somewhat unreasonable. For example, a special graph can be constructed, with the same features for nodes and different features for edges. If we consider edge features as weights and all weights of each node sum as one, it is interesting to find that no matter how the node features update, it will remain unchanged during the learning process.

The above phenomenon shows a fact, which has never been discussed in recent researches, that different graphs may have different preferences for node and edge features. To those graphs that edges possess a great impact, it is infelicitous to treat edge features as weights or labels. However, all existing works have ignored such a key fact. Our work's motivation is not only to integrate edge features into GATs but also to propose general models that spontaneously handle such preferences. To our best knowledge, we are the first to try to solve such problems. It should be emphasized that we do not want to present disparate models against state-of-the-art approaches. Since the attention mechanism has proved itself in lots of tasks, it is unnecessary to propose a completely new one. Our work is an extension of GATs\cite{velivckovic2017graph}, and all the improvements we made are served for our motivation.

\section{The proposed model}

\subsection{EGAT layer overview}

\begin{figure}[tbp] 
\centering
\subfigure{
\begin{minipage}[t]{0.48\textwidth}
\centering
\includegraphics[width=6.3cm]{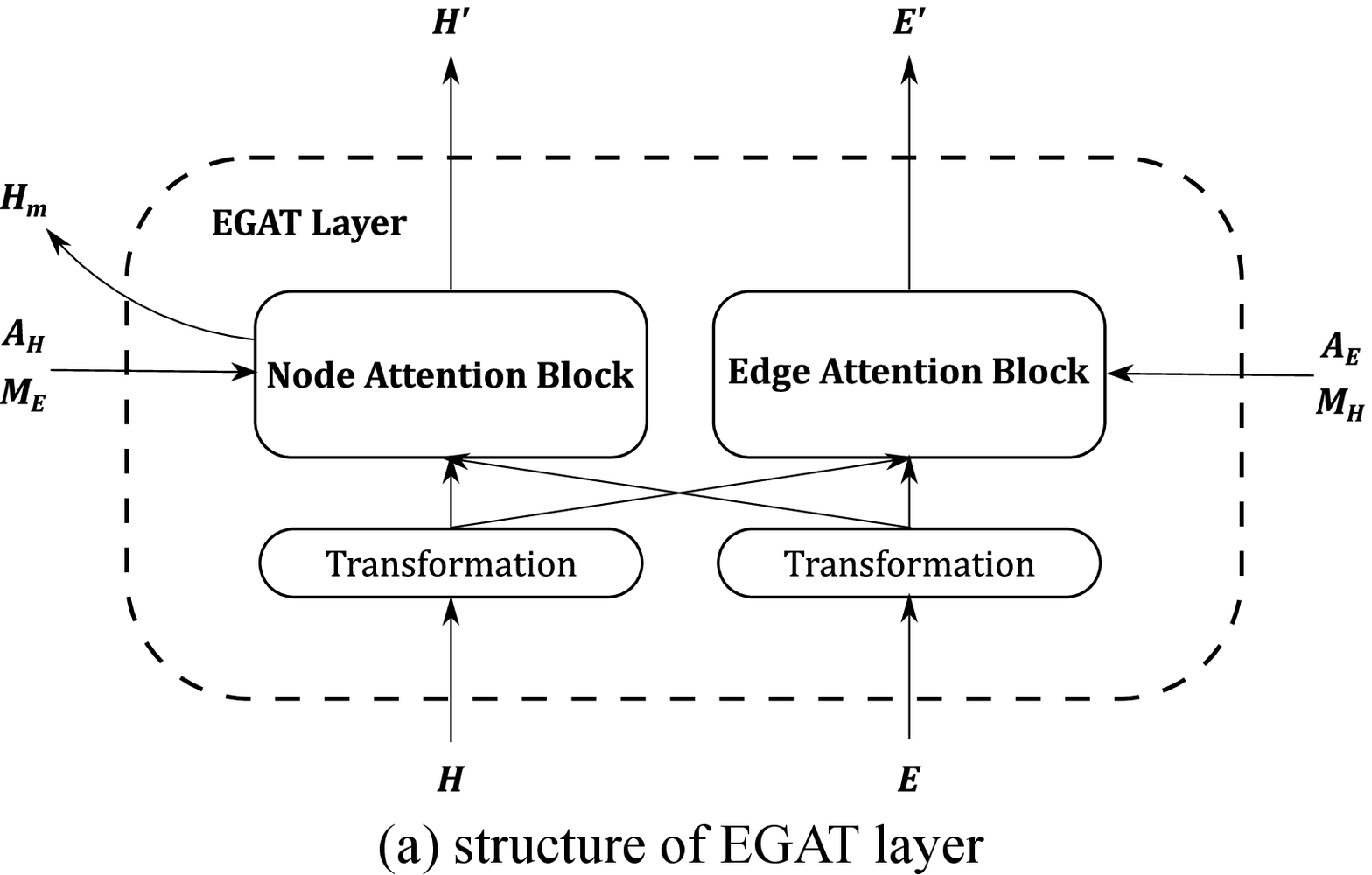}
\end{minipage}
}
\subfigure{
\begin{minipage}[t]{0.48\textwidth}
\centering
\includegraphics[width=6.3cm]{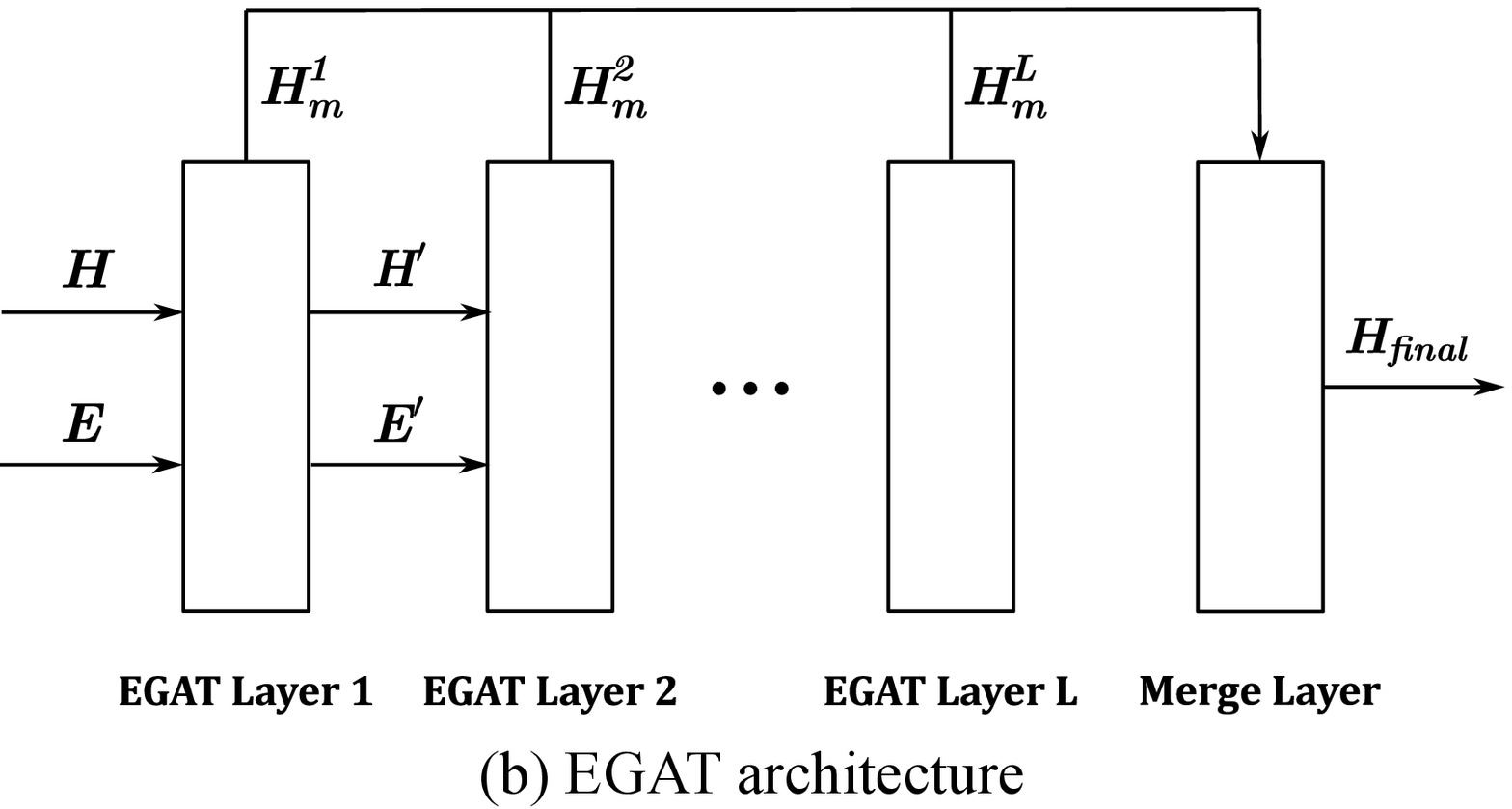}
\end{minipage}
}
\caption{(a) An illustration of one EGAT layer. It accepts $\textit{H}$ and $\textit{E}$ as inputs, and produces two sets of new features. $\textit{A}_H$ and $\textit{A}_E$ are adjacency matrices, while $\textit{M}_H$ and $\textit{M}_E$ are mapping matrices for nodes and edges, respectively. The $\textit{H}_m$ is edge-integrated node features generated by node attention block, which will only be used in the merge layer; (b) The architecture of EGATs. The model is constructed by several EGAT layers and a merge layer. The node and edge features generated from each iteration will be concatenated in the merge layer to achieve a multi-scale feature fusion. For convenience, the adjacency and mapping matrices are not shown in this figure, as well as the multi-head attention.}
\label{fig:fig1}
\end{figure}

A single EGAT layer contains two different blocks: node attention block and edge attention block. Each EGAT layer is designed in a symmetrical scheme; thus, the node and edge features can update themselves in a parallel and equivalent way. Figure ~\ref{fig:fig1} (a) gives an illustration of the EGAT layer. 

Each EGAT layer accepts a set of node features, $\textbf{H} = \{\vec{h}_1, \vec{h}_2 \dots, \vec{h}_N \}$, $\vec{h}_i \in \mathbb{R}^{F_H}$, as well as a set of edge features, $\textbf{E} = \{\vec{e}_1, \vec{e}_2 \dots, \vec{e}_M \}$, $\vec{e}_p \in \mathbb{R}^{F_E}$, as inputs. $N$ and $M$ represent the number of nodes and edges, while $F_H$ and $F_E$ symbolize the number of their respective features. After processing, the layer will produce high-level outputs, which include a new set of node features, $\textbf{H}' = \{\vec{h}_{1}', \vec{h}_{2}' \dots, \vec{h}_{N}' \}$, $\vec{h}_{i}'$ $\in \mathbb{R}^{F_{H}'}$, and a new set of edge features, $\textbf{E}' = \{\vec{e}_{1}', \vec{e}_{2}' \dots, \vec{e}_{M}' \}$, $\vec{e}_{p}'$ $\in \mathbb{R}^{F_{E}'}$. 

The cardinality of $F$ and $F'$ may be different (whether the $F$ is $F_H$ or $F_E$), since the linear transformations performed on the node and edge features are not same. We use two learnable matrices, $\textbf{W}_H \in \mathbb{R}^{F_H \times F_H'}$, and $\textbf{W}_E \in \mathbb{R}^{F_E \times F_E'}$, to achieve such transformations. For each node $i$ and edge $p$, their transformed features can be computed by $ \vec{h}_i^* = \mathbf{W}_H\vec{h}_i$, and $\vec{e}_p^* = \mathbf{W}_E\vec{e}_p$, respectively. Then, both of them will be fed into the node attention block and the edge attention block, which individually producing the new sets of node and edge features. Moreover, the adjacency and mapping matrices of nodes and edges will also be injected into the two blocks for ancillary computation. For simplicity, we will \textbf{re-use} some \emph{symbols}, which include $\textbf{H}$, $\textbf{E}$, $\vec{h}_i$, and $\vec{e}_p$. These symbols will have new meanings that characterize the features transformed by linear transformations in the rest of the paper. 

\subsection{Node attention block}
\label{sec:node_attention}
\begin{figure}[tbp] 
\centering
\subfigure{
\begin{minipage}[t]{0.48\textwidth}
\centering
\includegraphics[width=6.3cm]{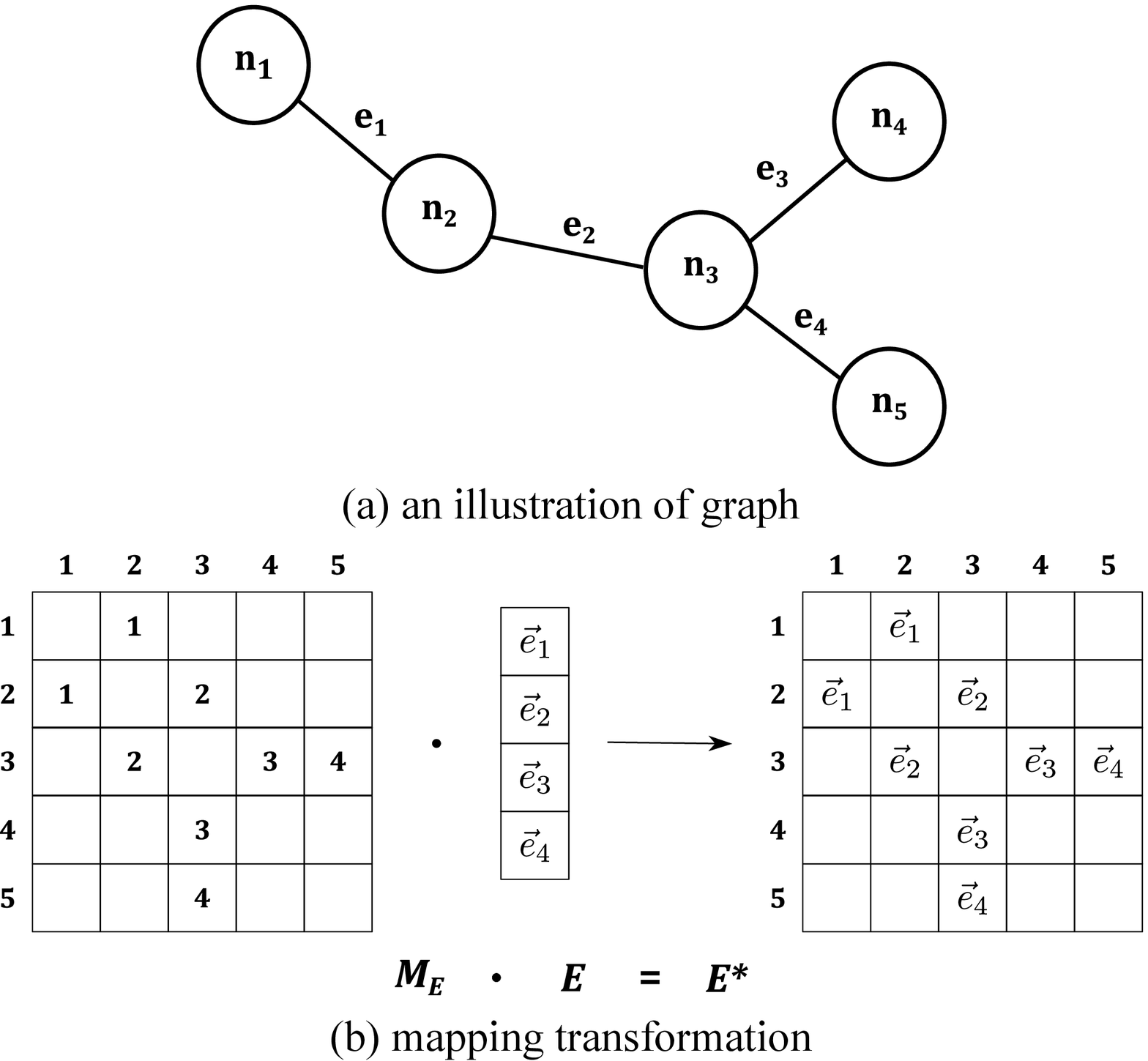}
\end{minipage}
}
\subfigure{
\begin{minipage}[t]{0.48\textwidth}
\centering
\includegraphics[width=6.3cm]{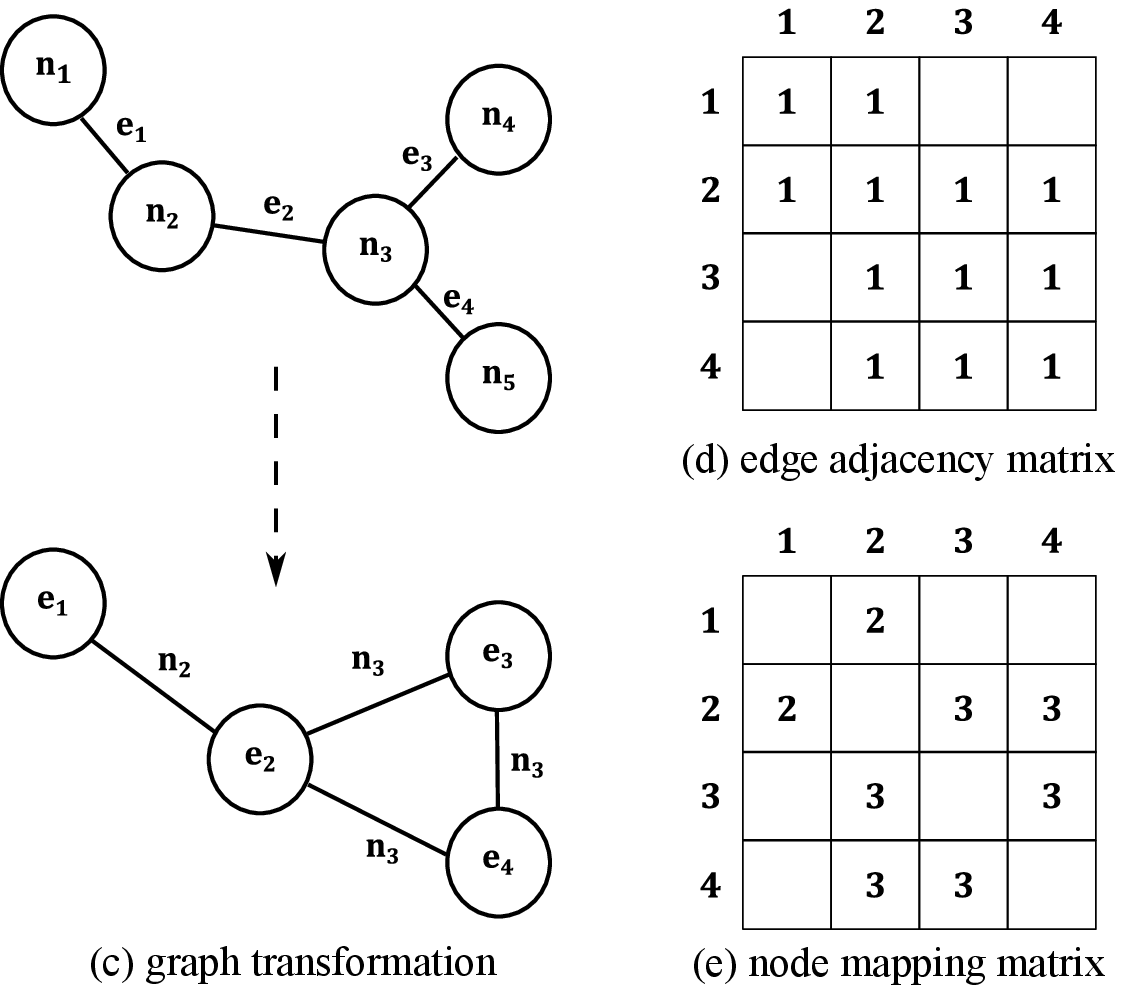}
\end{minipage}
}
\caption{\textbf{Left}: (a) An illustration of a regular graph; (b) An example of the mapping transformation performed on (a). ${M}_E$ is the edge mapping matrix in size of ${N} \times {N} \times {M}$, with its last dimension encoded in a one-hot scheme. For simplicity, we draw the matrix by replacing one-hot encoding vectors with non-zero indices. $E$ is the edge feature set with a size of ${M} \times {F}_E'$. ${M}_E$ should first be reshaped to ${N}^2 \times {M}$, and recover to ${N} \times {N} \times {F}_E'$ after multiplication. \textbf{Right}: (c) An example of graph transformation. The nodes and edges' roles are inversed in the new graph; (d) The adjacency matrix ${A}_E$ of the new graph, namely edge adjacency matrix. An identity matrix has been added to the matrix to pre-build the self-adjacent relations; (e) The mapping matrix ${M}_H$ of the new graph.}
\label{fig:fig2}
\end{figure}

The node attention block accepts $\textbf{H}$, a set of node features, and $\textbf{E}$, a set of edge features, and produces $\textbf{H}'$, a new set of node features. In $\textbf{E}$, the edge features are ranked in a preset order, so it is hard to find the relations between edges and their adjacent nodes. Thus, a mapping transformation will be first applied to $\textbf{E}$ in the block, to re-organize it into another common form $\textbf{E}^*$. Every element in $\textbf{E}^*$ can be represented as $\vec{e}_{ij}$, while $i$ and $j$ denote the nodes on each end of an edge. The transformation from $\textbf{E}$ to $\textbf{E}^*$ can be realized by matrix multiplication using an edge mapping matrix $\textbf{M}_E$, which is an $N \times N \times M$ tensor. Compared with the adjacency matrix, it expands the third dimension to indicate where each edge should be placed. Figure ~\ref{fig:fig2} (b) gives a simple example about the mapping process.

Before the multiplication, the edge mapping matrix should first be reshaped into $N^2 \times M$ so that $\textbf{M}_E$ and $\textbf{E}$ could have the same dimension of 2. Eventually, the multiplication result needs to reshape back with a size of $N \times N \times F_E'$, transforming the edge set $\textbf{E}$ into the adjacency form.  The edge mapping matrix is unique for a particular graph structure with determining orders of nodes and edges so that it can be constructed in a pre-processing step before the learning process.

Thanks to the adjacent form, the model can quickly seek out the edge between two specified nodes. Based on that, an edge-integrated attention mechanism can be performed on each node, generating the attention weights of its neighbors includes not only the features of the two nodes but also the edge connecting them. For each node $i$, the weight $w_{ij}$ will be computed for every $j \in \mathcal{N}_i$, where the $\mathcal{N}_i$ is the set including the first-order neighbors of node $i$ as well as the node $i$ itself. During the process, features will be concatenated, parameterized by a weight vector $\vec{a}$, and applied LeakyReLU as the activation function. Normalization will also be performed on these weights across all choices of node $j$, where $j \in \mathcal{N}_i$, by using a $\emph{softmax}$ function. The whole process can be formulated as follows:
\begin{equation}
   \alpha_{ij} = \frac{{\rm exp}({\rm LeakyReLU}(\vec{\textbf{a}}^T[\vec{h}_i \Vert \vec{h}_j \Vert \vec{e}_{ij}]))}{\sum_{k \in \mathcal{N}_i}{\rm exp}({\rm LeakyReLU}(\vec{\textbf{a}}^T[\vec{h}_i \Vert \vec{h}_k \Vert \vec{e}_{ik}]))}
\end{equation} 

It is interesting to note that, for each node, the aggregated features include not only the neighbors' but also the ones of itself. Without edge features, the problem can be solved by adding an identity matrix to the adjacency matrix. However, the introduction of edge features makes it more difficult. In our work, we use a tricky method by adding virtual featured self-loops to the graph. If a node does not have an edge that connected itself, a virtual self-loop will be attached to it. In particular, for every virtual self-loop, its features will be computed as an average of all its adjacent edges' features in each dimension as a compromise. All these operations should be done before being fed to the model.

After acquiring the normalized attention weights for each neighborhood, we can perform a weighted sum on these neighbor node features. In addition, a non-linearity $\sigma$ will be applied to these summation results. The final results, which is also the outputs of this node attention block, can be expressed as:
\begin{equation}
   \vec{h}_i' = \sigma(\sum\limits_{j \in \mathcal{N}_i}\alpha_{ij}\vec{h}_j)
\end{equation}

It should be noticed that we only aggregate the node features to generate the new set of node features. The edge features only play a part in weight computing but not a part of the new node features. It is for the clarity and symmetry of the model that we design such a strategy. If we merge edge features into nodes in each iteration, all the features may tangle up together and make the network more complicated and confusing. In fact, we also produce the set of edge-integrated node features $\textbf{H}_{m}$ in the node attention block. For each node $i$, we generate its new edge-integrated features as follows:
\begin{equation}
   \vec{m}_i = \sigma(\sum\limits_{j \in \mathcal{N}_i}\alpha_{ij}(\vec{h}_j \Vert \vec{e}_{ij}))
\end{equation}
However, these features will only be used in the last-level merge layer to achieve a multi-scale concatenation. They will never be passed to the next EGAT layer as the inputs.

\subsection{Edge Attention Block}
The node features can update themselves periodically in node attention blocks to acquire high-level features, whereas it is unreasonable to reuse the original low-level edge features during the weight computation. Besides, we also need high-level edge features to keep a balance of importance between nodes and edges. Thus, we proposed edge attention blocks, each of which accepts a set of node features, $\textbf{H}$, and a set of edge features, $\textbf{E}$, and produces $\textbf{E}'$, a new set of edge features.

A natural idea to realize such blocks is to update each edge's features by aggregating adjacent edges' features. In undirected graphs, we consider two edges are adjacent only if two edges have at least one common vertex. To achieve the aggregation, we adopt a tricky approach in our work by first switching the roles of nodes and edges in the graph. A similar concept on directed graphs has been proposed by Chen et al.\cite{chen2017supervised} for community detection. To achieve this, we create a new graph based on the original graph, whose nodes and edges are the edges and nodes of the original one, respectively. The transformation of the graph and the matrices structured by us are illustrated in Figure ~\ref{fig:fig2} (right).

The inputs of node and edge features are organized in the same sequential form. Thanks to the symmetric design, we can easily perform the attention mechanism on the new graph because the node feature set can be converted into the adjacency form by using $\textbf{M}_H$, the node mapping matrix, with no difficulty. For each edge $p$, the normalized attention weight of edge $q$ can be expressed as:

\begin{equation}
    \beta_{pq} = \frac{{\rm exp}({\rm LeakyReLU}(\vec{\textbf{b}}^T[\vec{e}_p \Vert \vec{e}_q \Vert \vec{h}_{pq}]))}{\sum_{k \in \mathcal{N}_p}{\rm exp}({\rm LeakyReLU}(\vec{\textbf{b}}^T[\vec{e}_p \Vert \vec{e}_k \Vert \vec{h}_{pk}]))}
\end{equation}
where $\mathcal{N}_p$ is the first-order neighbor set of edge $p$ (including $p$), and $\vec{\textbf{b}}$ is a weight vector with a size of  $\mathbb{R}^{2F_{E}' + F_{H}'}$. One noteworthy point is that, when we compute the attention weight of an arbitrary edge and the edge itself, it is no middle node between the two edges. In our experiments, we logically create an empty node between the two edges, by padding all the features of this virtual node as zeros. Like node features, the computing of new set of edge features can be represented as:

\begin{equation}
   \vec{e}_p' = \sigma(\sum\limits_{q \in \mathcal{N}_p}\beta_{pq}\vec{e}_q)
\end{equation}

\subsection{EGAT Architecture}
In this section, we present EGATs, which are constructed by stacking several EGAT layers and appending a merge layer at the tail. The architecture of EGATs is illustrated by Figure ~\ref{fig:fig1} (b).

Multi-scale strategies have been widely used to aggregate hierarchical feature maps in CNN models. Xu et al.\cite{xu2018representation} first introduced such strategies into GNNs and proposed jumping knowledge networks, which further improve the accuracy. Inspired by such works, we adopt a multi-scale merge strategy by adding a merge layer in EGATs. Unlike jump knowledge networks, we collect not only node features but also edge features. Edge features will be integrated into nodes in each EGAT layer, result in $\textbf{H}_m$, which we mentioned in ~\ref{sec:node_attention}. All $\textbf{H}_m$ generated from different iterations will aggregate together using a concatenation operation. Besides, we adopt the \emph{multi-head attention} in the merge layer to further stabilize the attention mechanism. Unlike GATs\cite{velivckovic2017graph}, our multi-head attention is performed on the unity of all EGAT layers rather than a single layer. $K$ independent multi-scale edge-integrated features would be computed and merged, resulting in the feature representation as follows:

\begin{equation}
   \vec{h}_i^{*} = \bigparallel_{k=1}^K (\bigparallel_{l=1}^Lm_i^{l,k})
   \label{equation:e2}
\end{equation}

where $L$ indicates the number of the EGAT layers, and $m_i^{l,k}$ represents the edge-integrated node features of node $i$ produced in iteration $l$ of the group $k$. To obtain a more refined representation, we apply a one-dimensional convolution to the results as a linear transformation and a non-linearity. For node classification tasks, a softmax function will be applied in the end to generate predicted labels. 

\section{Experiments}

We experientially assess the efficiency of EGAT by performing comparative evaluation against state-of-the-art approaches on several node classification datasets, which include both node-sensitive and edge-sensitive graphs. Besides, some additional analyses are also included in this section.

\subsection{Datasets}
We conduct our experiments on five node classification tasks containing both node-sensitive and edge-sensitive datasets. The former are graphs whose node features highly correlate with node labels, while the latter are those whose edges possess a dominant position. Such a division is somewhat relative, and it does not mean the features in a weak status have no contributions to the final results. 

\textbf{Node-Sensitive Graphs.} Three real-world node classification datasets, which include \emph{Cora}, \emph{Citeseer}, and \emph{Pubmed}\cite{sen2008collective}, are utilized in our experiments for node-sensitive graph learning. Such datasets are citation networks and have been widely used in graph learning research works as standard benchmarks. Notably, these datasets are undirected and do not have edge features within the graphs. For a fair comparison, in our work, we adopt the same dataset splits used in papers of GCNs\cite{kipf2016semi} and GATs\cite{velivckovic2017graph}.

\textbf{Edge-Sensitive Graphs.} We derive two trading networks, \emph{Trade-B} and \emph{Trade-M}, to test the effectiveness of our models on edge-sensitive graphs. The two datasets are financial-collaborative and refer to real-world trading records. For confidentiality, we cleaned and extracted some distinct patterns of abnormal behaviors from the original data provided by a bank and regenerated them as new datasets. In these datasets, each node represents a customer, with an attribute indicating the risk level of it. The edges, however, represent the relations among customers, whose features contain the number and total amount of recent transactions. \emph{Trade-B} is a binary classification dataset, which possesses 3907 nodes (97 of them are labeled) and 4394 edges. \emph{Trade-M}, however, is ternary classified, with 4431 nodes (139 of them are labeled) and 4900 edges. For both datasets, we separated the labeled nodes into three parts, for training, validation, and test, with a ratio of 3:1:1. The two datasets are directed initially; however, to make it suitable for EGATs, we converted them into an undirected form. 

\subsection{Experimental Setup}

For all the experiments, we implement EGATs based on the Pytorch framework\cite{paszke2019pytorch}. Because of the large memory usage for both adjacency and mapping matrices, we convert them into sparse forms to reduce the memory requirement and computational complexity during the learning process. The experimental setup for node-sensitive and edge-sensitive graphs are described as follows.

\textbf{Node-Sensitive Graphs.} Because all three citation networks do not possess even an edge feature, we generate one weak topological feature for each edge,  by enumerating the numbers of its adjacent edges. In those experiments, we adopt an EGAT model with $L$ = 2 and $K$ = 8, where $L$ and $K$ represent the number of the EGAT layers and the attention heads. For simplicity, we use the same numbers of the output features for every EGAT layer, where $F_H'$ = 8 and $F_E'$ = 4, for nodes and edges separately. A one-dimensional convolution operation are performed in the merge layer to produce C features (where C is the number of classes), followed by a softmax function. To improve accuracy, some techniques like dropout\cite{srivastava2014dropout} and $L_2$ regularization are also used in EGATs. All these experiments, but \emph{Pubmed}, were run on a machine with two GPUs of Geforce RTX 1080 Ti. Because of a larger requirement on video memory, \emph{Pubmed} was run on Tesla V100 instead.

\textbf{Edge-Sensitive Graphs.} \emph{Trade-B} and \emph{Trade-M} are the two edge-sensitive benchmarks used in our experiments. As we mentioned above, some virtual self-loops are added to the graphs before the training. For those datasets, we apply an EGAT model whose $L$ = 2 and $K$ = 8, with the same output feature dimension for each EGAT layer. Differing from the node experiments, we use three kinds of combinations of $F_H'$ and $F_E'$ here with different ratios, which can be listed as 8:4, 6:6, 4:8, respectively. Besides, all other details of this model are similar to those used for node-sensitive graph learning. All these experiments were run on a machine with two GPUs of Geforce RTX 1080 Ti.

\subsection{Results}
For the node-sensitive tasks, we report the classification accuracy on the test nodes after 10 runs, which are listed in Table ~\ref{tab:node-sensitive-results}. We compare our results against several strong baselines and state-of-the-art approaches proposed in previous works. In particular, we re-implement a two-layer GAT model\cite{velivckovic2017graph} by PyTorch, namely SP-GAT*, with $F'$, the number of hidden units, equals to 8. For a fair comparison, SP-GAT* accepts the same sparse representations of matrices used in our model.

The results show that EGATs are highly competitive against the state-of-the-art models on such node-sensitive graphs. We notice that there is a slight decrease for both \emph{Cora} and \emph{Citeseer} compared with SP-GAT*, which may be caused by the introduction of edge features. Since we generate the feature for each edge with the number of its adjacent edges, some interference may occur if these features are kind of useless. However, those negative effects are quite insignificant. Thanks to the symmetrical design, EGATs can adjust themselves during the learning and put more concentration on these useful features and produce acceptable results. In a word, EGATs can achieve high accuracy in node-sensitive classification tasks, surpassing the performance of most state-of-the-art approaches. 

\begin{table}
  \caption{Summary of the results on node classification accuracy, for Cora, Citeseer and Pubmed. SP-GAT* corresponds to the best result of GAT implemented by us with a sparse form.}
  \label{tab:node-sensitive-results}
  \begin{tabularx}{\columnwidth}{X X<{\centering} X<{\centering}X<{\centering}}
    \toprule
    \textbf{Method} & \textbf{Cora} & \textbf{Citeseer} & \textbf{Pubmed}\\
    \midrule
    MLP & 55.1\% & 46.5\% & 71.4\% \\
    ManiReg\cite{belkin2006manifold} & 59.5\% & 60.1\% & 70.7\% \\
    SemeiEmb\cite{weston2012deep} & 59.0\% & 59.6\% & 71.7\%\\
    LP\cite{zhu2003semi} & 68.0\% & 45.3\% & 63.0\%\\
    DeepWalk\cite{perozzi2014deepwalk} & 67.2\% & 43.2\% & 65.3\% \\
    ICA\cite{lu2003link} & 75.1\% & 69.1\% & 73.9\% \\
    Planetoid\cite{yang2016revisiting} & 75.7\% & 64.7\% & 77.2\% \\
    Chebyshev\cite{defferrard2016convolutional} & 81.2\% & 69.8\% & 74.4\% \\
    GCN\cite{kipf2016semi} & 81.5\% & 70.3\% & \textbf{79.0\%} \\
    Monet\cite{monti2017geometric} & 81.7\% & - & 78.8\% \\
    \midrule
    SP-GAT* & \textbf{82.5}$\pm$0.4\% & \textbf{70.8}$\pm$0.5\% & 78.1$\pm$0.4\% \\
    \textbf{EGAT} (ours) & 82.1$\pm$0.7\% & 70.3$\pm$0.5\% & 78.1$\pm$0.4\% \\
    \bottomrule
\end{tabularx}
\end{table}

\begin{table}
  \caption{Summary of the results on node classification accuracy, for Trade-B and Trade-M. The hyper-parameters $h$ and $e$ represent $F_H'$ and $F_E'$ used in our EGAT model, respectively.}
  \label{tab:edge-sensitive-results}
  \begin{tabularx}{\columnwidth}{X X<{\centering} X<{\centering}}
    \toprule
    \textbf{Method} & \textbf{Trade-B} & \textbf{Trade-M}\\
    \midrule
    SP-GAT* & 65.0\% & 46.4\% \\
    SP-GAT-sum* & 85.0\% & 51.1\% \\
    SP-GAT-avg* & 78.0\% & 71.4\% \\
    SP-GAT-max* & 81.5\% & 65.7\% \\
    \midrule
    \textbf{EGAT} ($h$ = 8, $e$ = 4)  & 87.5\% & 84.3\% \\
    \textbf{EGAT} ($h$ = 6, $e$ = 6) & 88.0\% & \textbf{85.4\%} \\
    \textbf{EGAT} ($h$ = 4, $e$ = 8) & \textbf{92.0\%} & 78.2\% \\
    \bottomrule
\end{tabularx}
\end{table}

For the edge-sensitive tasks, we report the mean classification accuracy on test nodes after 10 runs, and apply SP-GAT* and its variants to the benchmarks as comparisons. For SP-GAT*, we only feed the original node features as inputs. To ensure fairness, we further create three variants of SP-GAT*, by aggregating edge features into nodes as node features in advance using different functions, including sum, average, and max pooling. Besides, we evaluate the accuracy by comparing three EGATs with different ratios of $F_H'$ and $F_E'$. The comparative results are listed in Table ~\ref{tab:edge-sensitive-results}. EGATs show an incredible performance from the table, which is streets ahead of other approaches on both two datasets. For \emph{Trade-B} and \emph{Trade-M}, the best classification accuracy of EGATs can reach 92.0\% and 85.4\%. It is also interesting to observe that different datasets may possess different characteristics. For example, the edge features within \emph{Trade-B} can be better expressed by summing up together. On the contrary, the average operation may be more applicable to representing the edge features in \emph{Trade-M}. Despite their traits, EGATs can achieve high accuracy against these baselines on all these datasets, which means that EGATs can learn these characteristics of graphs spontaneously. To our best knowledge, there are no existing approaches that can process these kinds of graphs effectively.  

We also investigate the effects of the ratio of $F_H'$ and $F_E'$ on accuracy. According to the results, if the edge features may play a more important role than node ones, we recommend choosing a small or balance value of $F_H':F_E'$ so that edges will have a higher chance to show themselves. However, there may be exceptions in some cases. For example, the accuracy decreased to 78.2\% when we select a high $F_E'$ in \emph{Trade-M}. Due to the mutual effect of the features of nodes and edges, the model becomes complex, and it is hard to consider both features separately. So, if better performance is demanded, it is better to adjust these hyper-parameters several times to choose the most suitable ones.

\subsection{Complexity Analysis}
The complexity analysis of EGATs is given in this subsection. Since the constructions of adjacency and mapping matrices occur in a pre-processing step rather than the critical path, we merely ignore them and concentrate on the learning process. In EGATs, the matrix multiplication is the most time-consuming operation, which can be regarded as the entry point. Assume that now we have a graph with $N$ nodes and $E$ edges. In each node attention block, we introduce the edge features by applying a mapping transformation, which is actually matrix multiplication. It can be easily proved that the computation complexity of multiplying an $m \times n$ sparse matrix $A$ and an $n \times p$ dense matrix $B$ can be reduced to $O(cp)$, where $c$ denotes to the number of non-zero elements in $A$. Thus, the complexity of such a transformation is $O(E)$, for the reason that $p$, the number of features, can be seen as a constant. Because the complexity of GATs is no less than $\textit{O(E)}$ in each iteration, so the introduction of edges in node attention blocks will not significantly increase the complexity.

Things may be a little different occurring in edge attention blocks. Consider a graph with one central node and $K$ neighbors. Because every two edges are neighbors, when we switch the roles of nodes and edges, the number of edges in the new graph, $E^*$, is on the order of $O(K^2)$. When we extend this conclusion to a regular graph with $N$ nodes and $E$ edges, the number of non-zero elements in the converted mapping matrix can be represented as $O(\sum_{i=1}^{N}{d_i^2})$, where $d_i$ indicate the degree of each node in the graph. Thus, the complexity of edge attention block is in the same order. However, based on our experimental results, the delay of EGATs is quite acceptable on the benchmarks and those graphs with a similar scale. Besides, all the test datasets except \emph{Pubmed} can run on Geforce RTX 1080 Ti without exceeding the memory limit. If someone has higher performance requirements, some modifications could be made in edge attention block. For example, each edge can regard the two adjacent nodes as virtual edges and only aggregate the edge part of $\textbf{H}_{m}$ during the learning.

\section{Conclusions}

We proposed edge-featured graph attention networks (EGATs), novel edge-integrated graph neural networks that performed on graphs with node and edge features. We incorporate edge features into GNNs and present a symmetrical approach to exploit them. To our best knowledge, we are the first to incorporate edges as node-equivalent entities, point out graphs' different preferences for features, and handle them spontaneously. The results demonstrate that EGATs have successfully achieved state-of-the-art performance on node classification tasks, especially for edge-sensitive datasets.

There are some potential improvements to EGATs that could be addressed as future work. In EGATs, we update each edge's features by aggregating its neighbors' information. However, the number of neighbors of each edge may be huge. Despite transforming matrices into sparse forms, the models still need a large memory usage when it performed on large-scale graphs. Thus, it is better to find an improved way to reduce the models' memory requirement. Besides, EGATs do not naturally support directed graphs as well as multi-graphs. We intend to achieve these extensions further on.

\section*{Broader Impact}

In this work, edge-featured graph attention networks (EGATs), novel edge-integrated graph neural networks, were proposed to perform node classification on those graphs with node and edge features. This work has the following potential positive impact on society. First, the models proposed in this paper are kind of versatile and have a broad application foreground in many fields. For example, the models can be applied in the financial sector, as an aid to finding those people who may have a suspicion in financial fraud, money laundering, etc. Second, given the absence of edge features in traditional GNN approaches, our work may attract the attention of other researchers, spawning a series of related research, to enhance further the basic framework of graph neural networks from a theoretical level. At the same time, this works may have some negative consequences with a small probability. Because there are very few works trying to exploit edge features in graph neural networks, this field is still kind of immature and receives little attention. Therefore, the negative impact of our models on society are quite unclear and need further exploration. Besides, we should be cautious about the result of the failure of the system. It should be noticed that the prediction results of our models should only be regarded as an auxiliary reference rather than a definite truth. Users of the models should perform a second manual verification by their own to ensure the authenticity of the results. We will not be responsible for the negative effects caused by the wrong prediction of EGATs.

\end{document}